\documentclass[letterpaper]{article} % DO NOT CHANGE THIS
\usepackage[submission]{aaai23}  % DO NOT CHANGE THIS
\usepackage{times}  % DO NOT CHANGE THIS
\usepackage{helvet}  % DO NOT CHANGE THIS
\usepackage{courier}  % DO NOT CHANGE THIS
\usepackage[hyphens]{url}  % DO NOT CHANGE THIS
\usepackage{graphicx} % DO NOT CHANGE THIS
\usepackage{natbib}  % DO NOT CHANGE THIS AND DO NOT ADD ANY OPTIONS TO IT
\usepackage{caption} % DO NOT CHANGE THIS AND DO NOT ADD ANY OPTIONS TO IT
\usepackage{algorithm}
\usepackage{algorithmic}

\usepackage{amsmath}

\usepackage{newfloat}
\usepackage{listings}
\DeclareCaptionStyle{ruled}{labelfont=normalfont,labelsep=colon,strut=off} % DO NOT CHANGE THIS
\floatstyle{ruled}
\newfloat{listing}{tb}{lst}{}
\floatname{listing}{Listing}
\title{Systematic Bias in Sample Inference and its Effect on Machine Learning}

\author {
    Owen O'Neill,\textsuperscript{\rm 1}
    Fintan Costello, \textsuperscript{\rm 2}
}
\affiliations {
    \textsuperscript{\rm 1} University College Dublin\\
    \textsuperscript{\rm 2} University College Dublin\\
    owen.o-neill.1@ucdconnect.ie, fintan.costello@ucd.ie
}

\begin{document}

\maketitle

\begin{abstract}
A commonly observed pattern in machine learning models is an underprediction of the target feature, with the model’s predicted target rate for members of a given category typically being lower than the actual target rate for members of that category in the training set. This underprediction is usually larger for members of minority groups; while income level is underpredicted for both men and women in the ‘adult’ dataset, for example, the degree of underprediction is significantly higher for women (a minority in that dataset). We propose that this pattern of underprediction for minorities arises as a predictable consequence of statistical inference on small samples. When presented with a new individual for classification, an ML model performs inference not on the entire training set, but on a subset that is in some way similar to the new individual, with sizes of these subsets typically following a power law distribution so that most are small (and with these subsets being necessarily smaller for the minority group). We show that such inference on small samples is subject to systematic and directional statistical bias, and that this bias produces the observed patterns of underprediction seen in ML models. Analysing a standard sklearn decision tree model's predictions on a set of over $70$ subsets of the `adult' and COMPAS datasets, we found that a bias prediction measure based on small-sample inference had a significant positive correlations (0.56 and 0.85) with the observed underprediction rate for these subsets.
\end{abstract}

\section{Introduction}

Over the past decade, ML has been increasingly applied to several sensitive areas. Criminal justice, healthcare and banking all apply ML to inform their decisions, which can have significant impact on people's lives. Given the sensitive nature of these areas, and historical discrimination, it is vital to understand the sources of any biases exhibited by the model. 
In the literature, examples abound of standard ML approaches showing significant bias towards certain demographics \cite{corbettdavies2018, chakraborty2021, mehrabibias, turnerlee_2018, suresh2019framework, corbett2017, barocas2017fairness, hutchinson2019, friedler2019}. Two primary sources of bias have been identified: data bias and algorithmic bias. Data bias may be due to errors in data collection, non-representative or skewed samples or active prejudice within the problem area (e.g. societal gender/racial bias). It is assumed that biased data will result in biased predictions, assuming no correction methods are applied to the model. 
Algorithmic bias is bias introduced by the model and will result in biased predictions even when using `unbiased' data. The causes of algorithmic bias are more nebulous, most often attributed to some flaw in the algorithm's design or the inference process of ML itself. 
While attempts have been made to quantify bias in data \cite{ntoutsi2020bias, mehrabibias, olteanu, suresh}, algorithmic bias is primarily seen as a problem to be corrected rather than as a phenomena to be measured. However, even ignoring the challenges in defining fairness, modelling this bias would be a useful addition to real world ML application. Therefore, our research aims to examine ML inference, and the statistical processes underpinning it, in order to understand the patterns of bias seen in the literature.

We focus on a particular objective and quantifiable measure of bias proposed by \cite{cunningham2020}, based on the difference between the rate at which members of a given group have the target variable in the dataset (the `observed target rate') and the rate at which members of that group are predicted to have the target by an ML algorithm trained on the same dataset (the `predicted target rate'). This measure reveals a common pattern of `underprediction', where the predicted target rate for a given group is reliably lower than the observed target rate for that group in the dataset. This measure also reveals a related `underprediction bias' against minority groups (where the degree of underprediction is higher for the minority than the majority). An example of these patterns in the `adult’ dataset can be seen in tables 5-7, where the target rate is underpredicted for both men and women, but where underprediction for the minority (female) group is larger than for the majority (male) group ($25\%$ versus $17\%$).

\section{Background}
\subsection{Bias Metric}
Before we begin our analysis, we must elaborate on our bias metric. We refer to this metric as `underprediction bias' as we have frequently observed ML models underestimating the target rate in their predictions, relative to the observed sample rate. We are interested in the rate at which failures of prediction occur for the minority and majority groups ($G=0,G=1$). We measure failures of prediction relative to the dataset, by comparing the number of minority/majority group members in the dataset who have the target feature ($T=1$) against the predictions of a given machine learning algorithm (which we write as $P(T=1 | X=1, G)$, where $X$ is the feature used to inform the prediction). For each group we normalise by dividing by the actual number with the target variable, so the degree of bias for a given group $i$ is given by
\[ U(X,i)= \frac{P(T=1|G=i) - P(T=1 | X=1, G=i)}{P(T=1|G=i)}\]
The larger this number, the higher the probability that a given member of group $i$ who should be predicted to be $T=1$ is actually predicted to be $T=0$, and the more members of that group are disadvantaged by the algorithm's predictions. A value of $U(X,i)=0.1$, for example, would indicate that the machine learning algorithm predicts the target variable for group $i$ at a rate $10\%$ lower than the rate of that variable in the dataset. 

\subsection{Distributional Inference from Samples}
Our investigation begins with inference at its most fundamental: given a sample, what can we infer about the population from which it was drawn? Consider a situation where we are given a sample of $N$ items drawn from some population, $K$ of which have a particular feature (which we'll call $A$). We want to make predictions about the probability of $A$ in the population, using only the given sample (and no information beyond that sample). 

A typical assumption is that the correct probability estimate for $A$ in the population, given the observed sample, is equal to the sample proportion:
\begin{eqnarray}
 \label{eq:prop}
 Pr(A) = \frac{K}{N}
\end{eqnarray}
However, this assumption is fundamentally incorrect. To see why, consider an extreme case, where you are shown a sample of $2$ items (neither of which are instances of $A$) that come from one population, and a sample of $20$ items (none of which are instances of $A$) that come from another population. The sample proportions in both cases are $Pr(A) = 0$. Concluding that $A$ has a probability of 0 in both populations is incorrect, the sample of size $2$ is far too small to justify such a statement. Proposing that $A$ has the same probability in both populations is also inaccurate. $P(A) = 0.25$ could reasonably hold in the first population (the probability of drawing a sample of 2 items neither of which are $A$, from a population where $P(A) = 0.25$, is $(1 - 0.25)^2 = 0.56$; a more than $50\%$ chance), but $P(A) = 0.25$ is extremely unlikely to hold in the second population (the probability of drawing a sample of 20 items, none of which are $A$, from a population where $P(A) = 0.25$, is $(1 - 0.25)^{20} = 0.003$; a less than $1\%$ chance). \newline

Therefore, an alternative approach is necessary. In order to determine the most likely population to have generated our sample, we base our inference on the distribution of all possible populations (`distributional inference' (DI)). Beginning with simulation, for a given sample size N we run the function PROBABILITIES(N) (see Algorithm 1). This function loops $10,000$ times, on each cycle randomly picking a value for the population probability p = P(A) of some event A (p is drawn uniformly from the range $0 \ldots 1$ inclusive). On each cycle the function SAMPLE(p, N) then draws a sample of N items from the population with $p = P(A)$, by randomly picking N values q, drawn uniformly from the range $0 \ldots 1$ inclusive: Cases where $q < p$ are counted as an instance of event A in our sample. SAMPLE(p, N) then returns the number of cases which were counted as an instance of event A in the drawn sample. For each K from 0 to N the function PROBABILITIES(N) has an associated storage list P$_K$: On each cycle of our simulation where the drawn sample contains K instances for event A, we add the probability $p = P(A)$, which generated that sample to the associated storage list $P_K$. Each list $P_K$ thus holds the set of population probabilities $p = P(A)$ that generated samples of N events containing K instances of A. After running this simulation for $10,000$ cycles, we then display the average generating probability that produced samples with $K = 0, K = 1,..., K = N$. This average generating probability represents the statistically optimal estimate for the underlying population probability $P(A)$, given an observed sample of size N that contains K instances of A. \newline

\begin{algorithm}[tb]
\label{alg:algorithm}
\caption{}
\begin{algorithmic} %[1] enables line numbers
\STATE \textbf{def} sample(p,N):
\STATE Let $k=0$.
\FOR{i in \textbf{range}(1, N+1)}
\STATE Let $q=$random.random()
\IF{$q>p$}
\STATE $k=k+1$
\ENDIF
\ENDFOR
\STATE \textbf{return} k

\STATE \textbf{def} probabilities(N):
\STATE Pk  =  [ [ ] \textbf{for} i in  \textbf{range}(N+1)]

\FOR{j in \textbf{range}(1, 10,000)}
\STATE Let $q=$random.random()
\STATE Let $K=$ \textbf{sample}(p,N)
\STATE $P_{k}$ [K].append (p)
\ENDFOR

\FOR{K in \textbf{range}(0, N+1)}
\STATE Pr = K/N
\STATE Pg = np.\textbf{mean}($P_{k}$ [K])
\ENDFOR
\STATE \textbf{return} K, Pr, Pg

\end{algorithmic}
\end{algorithm}

Table \ref{table:1} shows the output from this simulation for values $N = 16, 4$ compared with the sample proportions. It is clear from this table that, for a given sample of $N$ items containing $K$ instances of event $A$, the average probability $P$ that generated that sample differs from the sample proportion $Pr = K/N$. Specifically, the average generating probability (and so the normatively correct, optimal estimate for the population probability, given the sample in question) is regressive toward $0.5$ with the degree of regression increasing as the sample size $N$ falls. Therefore, for a sample containing a majority and minority group, different target rates will be predicted for each group, even if sample target rates are equal. 
These predictions actually follow a well-known result in epistemic probability theory, the Rule of Succession (RoS), given by

\begin{equation}
 \label{eq:ros}
 P(A) = \frac{K + 1}{N+2}
\end{equation}

This expression has been proved elsewhere in various ways, with the strongest and most general proof being given by \cite{definetti1937}. As \cite{zabell1989}, in a very interesting presentation of the history and various proofs of the RoS, notes, “[I]n order to attack [De Finitti’s proof] one must attack the formidable edifice of epistemic probability itself.” \newline

\begin{table}[h!]
	\centering
	\caption{Average Generating Probability P and Sample Proportion Pr = K/N for Samples of Size N Containing K Instances of Some Feature A, as Generated by Algorithm 1}
	\label{tab:probs}
	\begin{tabular}{c c c} 
		& N=16 &  \\ [0.5ex] 
		K & Pr & P \\ [0.5ex] 
		\hline
		0 & .00 & .06 \\ 
		4 & .25 & .28 \\
		8 & .50 & .50 \\
		12 & .75 & .72 \\
		16 & 1.00 & .94 \\ [1ex]  
	\end{tabular}
	\quad
	\begin{tabular}{c c c} 
		& N=4 &  \\
		K & Pr & P \\ [0.5ex] 
		\hline
		0 & .00 & .17 \\ 
		1 & .25 & .33 \\
		2 & .50 & .50 \\
		3 & .75 & .67 \\
		4 & 1.00 & .83 \\ [1ex]   
	\end{tabular}
	\label{table:1}
\end{table}

\subsection{The Beta Prior}

While the RoS was stated before the advent on Bayesian Statistics, the result is equivalent to assuming a uniform prior, $Beta(1,1)$, and updating this using the sample.

A central property of the Beta distribution is that, given that we have the prior distribution $p_A \sim \mathrm{Beta}(a,b)$ and have also observed $K$ occurrences of $A$ in a sample of $N$ events, then the updated or posterior distribution for $p_A$ will be the Beta distribution

\begin{equation} \label{eq:conjugate}
p_A \sim \mathrm{Beta}(a+K,b+N-K)
\end{equation} 

(a Beta prior necessarily gives a Beta posterior; the prior and the posterior are `conjugate', to use Bayesian terminology). Equation \ref{eq:conjugate} gives a probability distribution for the unknown generating probability $p_A$, given the observed sample $K/N$ (and the prior parameters $a$ and $b$). The expected value or mean of this Beta distribution is

\begin{equation}
\langle \mathrm{Beta}(a+K,b+N-K) \rangle = \frac{K+a}{N+a+b}
\end{equation}

This expression gives the expected value for our unknown generating probability given the observed sample, and so is the theoretically optimal estimate for that probability (given the sample, and given our priors $a,b$).

In our `inference from samples' task, we assume no information about the generating probability $p_A$ apart from that given by our sample: prior to seeing the sample, we would consider every possible value of $p_A$ as equally likely. For a Beta distribution with $a = b = 1$ we have 

\begin{equation} 
	P( p \leq p_A \leq p +\Delta p)  = \frac{p^0(1-p)^0}{B(1,1)}\,\Delta = \Delta p
\end{equation}

and every possible value of $p_A$ is equally likely (the chance of $p_A$ falling in a given $p \ldots p +\Delta p$ range is simply equal to the size of that range). The distribution $Beta(1,1)$ thus represents the uniform distribution or the `uninformative prior' for probability inference from samples, and so the expected value of our unknown generating probability $p_A$, given a sample of $N$ items of which $K$ are instances of $A$ (and no other information beyond that sample) is the RoS. Note that this effect is unavoidable, any attempts to reduce these regressive effects will deviate from DI. Other choices of prior will be explored in the discussion. 

\section{Machine Learning Inference}
\subsection{The Effect of Predictor Variables}

Our investigation thus far has only considered the bias arising from inference on data containing target and group membership. It is important to note, however, that it would be more accurate to describe these effects as occurring between target and predictor variable. Group membership is mentioned above to highlight the observed regression in the context of discrimination against minority groups, but the effects can be observed using any predictor variable. 

It is useful now to consider ML inference. ML is fundamentally a process of achieving the optimal inference from a real world sample. As we saw in the previous section, the theoretically optimal inference (DI) necessarily produces systematic bias. Therefore, we would expect ML algorithms to be subject to similar biases to those described above. Due to the highly sensitive scenarios where ML algorithms are being applied this is a cause for significant concern. That the DI approach systematically effects minority groups more than majority groups exacerbates the situation further.

In most ML data sets there are thousands of data points with many predictor variables under consideration, each with differential association with the target variable. The scale of this data appears to cast doubt over the impact of the bias we've described above as the `plus one over plus two' becomes negligible as $K$ and $N$ increase past $30$. However, a more realistic way to look at predictor variables in an ML context would be to consider predictor variable combinations (PVCs). Put simply, when presented with a new individual for classification, the ML system references a subset of similar individuals (i.e. individuals with a similar PVC). The inference is performed on this similar subset, not the dataset as a whole. Therefore, even with a large dataset, if there are few examples of the PVC to reference, the inference takes place on a small sample. To model ML we simply take one of these inference scenarios.

Consider a sample of size $100$, with three variables; target, group (majority or minority) and PVC (which we'll call $X$) membership. The total sample is described in Table \ref{tab:pv1}, with a breakdown for majority and minority in Table \ref{tab:pv2}.

As before, the target is rare. Therefore, an informative PVC would co-occur with the target variable at a high rate. For simplicity, we assume a `perfect predictor': X=0 when T=0 and X=1 when T=1. 

\begin{table}[!htb]
      \caption{Proportions of Sample used for Inference (N=100)}
      \centering
        \begin{tabular}{l|ll|ll}
                & X=0 & X=1 & Maj & Min \\ \hline
            Target=0          & 0.8 & 0 & 0.64 & 0.16 \\
            Target=1          & 0 & 0.2 & 0.16 & 0.04 \\ 
        \end{tabular}
\label{tab:pv1}
\end{table}

\begin{table}[!htb]
      \caption{Counts of Sample used for Inference (N=100)}
      \centering
        \begin{tabular}{l|ll|ll}
                & Maj &  & Min & \\
                & X=0 & X=1 & X=0 & X=1 \\ \hline
            Target=0          & 64 & 0 & 16 & 0 \\
            Target=1          & 0 & 16 & 0 & 4 \\ 
        \end{tabular}
\label{tab:pv2}
\end{table}

We assume inference takes place with no information beyond that given in the sample, and so apply the RoS to find the most likely generating probabilities for the target variable. The predicted conditional probabilities are:
\begin{equation}
\begin{aligned}
P(Target=1 | X=1, Maj) &= \frac{16+1}{(0+16)+2} = 0.94 \\
P(Target=1 | X=1, Min) &= \frac{4+1}{(0+4)+2} = 0.83
\end{aligned}  
\end{equation}

(since in the majority group we have $N=16$ occurrences of the PVC, and $K=16$ of these occur with the target, while in the minority group we have $N=4$ occurrences of the PVC, and $K=4$ of these occur with the target).
Based on this sample, therefore, a normatively correct reasoner will infer that the probability of the target for a member of the majority with PVC $X$ is $0.95$, while the probability of the target for a member of the minority with PVC $X$ is $0.83$. Both of these probabilities are less than the sample proportion ($Pr(Target|X)=1$), and so are underpredictions. Further, the degree of underprediction is greater for members of the minority than for the majority. This differential association is a `rational' (in a statistical sense) consequence of the fact that we are making inference from samples, and that the sample sizes for our two groups are different. Returning to our example, this implies that for an individual with characteristics indicating a high salary, being male makes them more likely to be predicted to have a high salary. 
 
Assuming that individuals are classified as having the target variables based only on the presence of PVC $X$, and recalling that in our example the PVC occurred with $20\%$ of both groups (because we assume the PVC co-occurs perfectly with the target variable in the sample), we see that the proportion of individuals in each group identified as having the target variable will be
\begin{equation}
\begin{aligned}
& \textrm{Prop. of majority predicted to have T=1} = 0.94*0.2 = 0.188 \\
& \textrm{Prop. of minority predicted to have T=1}  = 0.83*0.2 = 0.166
\end{aligned}
\end{equation}
Finally, multiplying these by the proportion of each group in the sample overall, we get estimated probabilities of association between group membership and target variable (Table \ref{tab:pv5}).

\begin{table}[]
\centering
\caption{Proportion table for DI from samples, with percentage rise or fall relative to sample proportions in Table \ref{tab:pv2}.}
\label{tab:pv5}
\begin{tabular}{l|ll}
           & Maj & Min \\ \hline
Target=0          & 0.65  & 0.17   \\
Target=1          & 0.15 (6\% fall) & 0.03 (25\% fall)   \\ 
\end{tabular}
\end{table}

Table \ref{tab:pv5} shows the resulting probabilities of target prediction for each category. These results show an underprediction for both groups (relative to the actual $20\%$ rate in each class in the sample), but a greater underprediction for the minority. Here we have demonstrated that even with unbiased data (majority and minority have equal target rates in the sample) we still see the same patterns of underprediction. This sample size bias is unavoidable, even in these idealised theoretical examples, without departing from DI. \newline

\subsection{Predictions of the Overall Sample}

To expand the above analysis to measure overall model bias on a dataset, we must consider the distribution of PVC sizes. The PVCs in many datasets roughly follow a power law distribution (demonstrations of this in `adult' and COMPAS can be seen in the supplementary material). A subset of these will be considered relevant by the model and used for inference. This subset (e.g. the PVCs in the leaves of a decision tree) also roughly follows a power law, implying that the vast majority of PVCs occur infrequently ($<100$ times).

Therefore, for an example of ML inference, we simply apply the single PVC case, described in section 3.1, distributed by a power law. This highlights the exact source of this type of bias in ML: the distribution of PVC sizes, which we refer to as `exponential spread' (ES):

\begin{equation}
    \label{eq:ES}
    ES = \sum_{i} \frac{p_{i}}{i}
\end{equation}

where $i$ ranges across all observed PVC sizes in our dataset and $p_i$ is the proportion of the dataset contained in a PVC of size $i$. More infrequent PVCs in a group will result in more predictions with high levels of bias, increasing the overall bias for that group. The minority group, being smaller, generally has more infrequent PVCs which leads to it experiencing more bias.

The subset of PVCs selected by the model tend to co-occur with the target at a high rate (assuming the target variable occurs at a low rate in the sample). To simplify, in our example we let $S>0.5$ be the average co-occurrence with the target for the PVCs selected by the model.

Therefore, for a given leaf containing F instances, the inferred probability will be 
\begin{equation}
\begin{split}
\frac{S*F+a}{F+2a}
\end{split}
\end{equation}

Suppose we have two groups (majority and minority) with the same combinations of features. Let $N$ be the size of the overall sample, $R$ the proportion of the sample in the minority group, $S_{1}$ and $S_{2}$ be the average co-occurrence with the target for the majority and minority groups respectively and $a=b=1$ for the Beta prior. The difference in predicted probability compared to the sample will be:

\begin{equation}
\begin{split}
b(F)= \frac{(\frac{S_{2}*F*R+1}{F*R+2} - S_{2})} {(\frac{S_{1}*F*(1-R)+1}{F*(1-R)+2} - S_{1})}
\end{split}
\end{equation}

The overall expression for the model's prediction bias will be:

\begin{equation}
\begin{split}
B = \int_1^{N} b(F) * P(F) dF
\end{split}
\end{equation}

where N is the size of the dataset (and so the largest possible leaf size) and $P(F)$ is the probability of a leaf of size $F$ occurring (since we assume a power law distribution, this will be $\frac{1}{F^X}$, where X is defined in the power law equation).

\begin{figure}[ht]
    \centering
    \caption{Underprediction for Various Power Law Distributions}
    \includegraphics[scale=0.5]{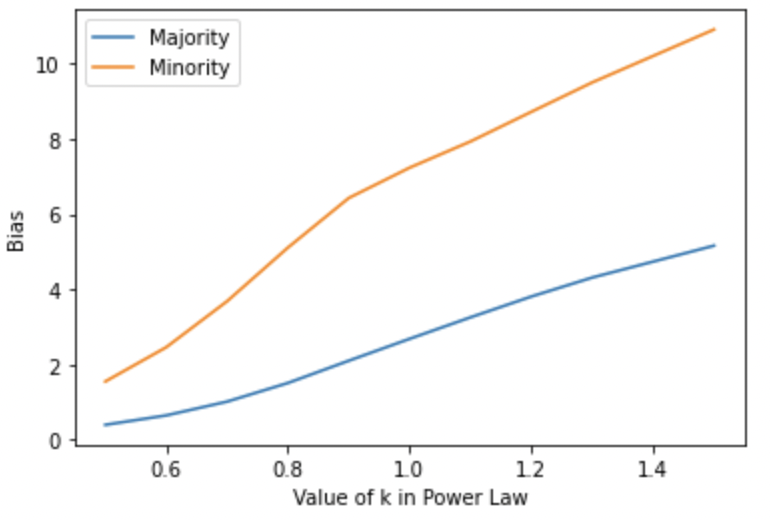}
    \label{fig:plaw}
\end{figure}

The underprediction seen for these power law distributions for each group can be seen in Fig. \ref{fig:plaw}. 

These graphs demonstrate the concepts we have seen in the simpler examples. Infrequent combinations will be more affected by the regressive effects of DI and so will be subject to more bias. This effect will be more pronounced in the minority group as it necessarily contains more infrequent combinations. Therefore, underprediction increases with the number of infrequent combinations.

\begin{figure}[ht]
    \centering
    \caption{Underprediction for Various PVC `Qualities'}
    \includegraphics[scale=0.5]{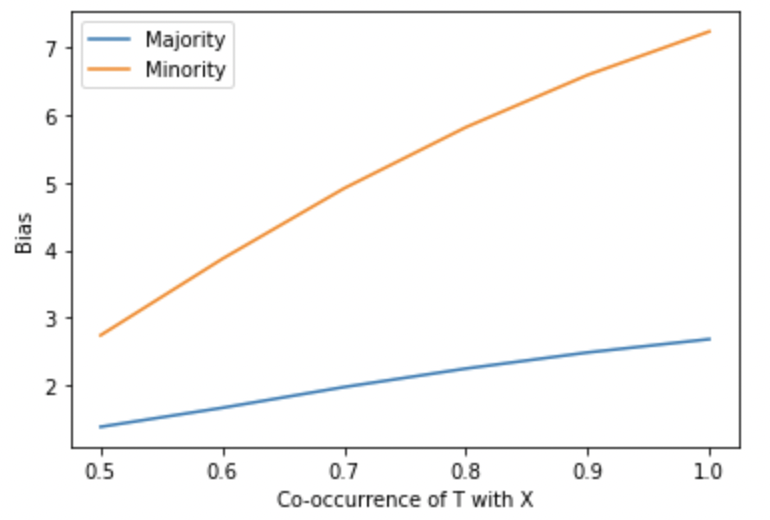}
    \label{fig:trate}
\end{figure}

Fig. \ref{fig:trate} reveals a somewhat paradoxical result. The greater the co-occurrence of the PVCs, i.e. the greater quality, the higher the underprediction. However, this is consistent with our previous examples. The effect of DI is regressive towards 0.5, with the highest regression at 1. Therefore, we see this pattern in our numerical example. This result has troubling implications for our inference; choosing better variables results in more bias.

\subsection{Decision Tree Bias}
The above analysis gives an expression for bias $B$ assuming that leaf size $F$ for leaves in a decision tree with $S>0.5$ (that is, where the target variable occurs at a rate above $50\%$) is distributed following a power law. Here we consider the number of leaves of size $F$ with $S > 0.5$ that we would expect to see in a particular group $G$ of size $N$, assuming that the target rate for members of that group has some value $p$. In this situation we can assume that target occurrences are distributed randomly across the set of leaves of size $F$, so that to a first approximation the probability of a leaf of size $F$ having $S>0.5$ is given by the complementary cumulative binomial 
\[1-Bin(F/2,F;p)\]
(the probability of getting a sample of size $F$ containing more than $50\%$ successes, when the probability of a single success is $p$) and, since the target variable is predicted only for leaves with $S>0.5$, this expression gives the expected predicted target rate for individuals in leaves of size $F$. Note that when $p<0.5$ the value of this expression necessarily falls with increasing $F$ (the binomial distribution becoming more peaked around $pF$ as $F$ increases), and that its value when $F=1$ is exactly $p$. This means that when $p<0.5$, the probability of a leaf of size $F$ having $S>0.5$ is less than $p$ for all $F>1$. If leaf size is distributed proportional to some power law with exponent $X$, the the total predicted target rate for individuals in group $G$ is
\[\sum_1^N \frac{1-Bin(F/2,F;p)}{F^X} \]
and, since this is simply an average of predicted target rate across all leaf sizes, this total predicted target rate is necessarily less than the group target rate $p$ when $p<0.5$ (with the difference falling with increasing exponent $X$). An analogous argument shows that this this total predicted target rate is necessarily greater than the group target rate $p$ when $p>0.5$ (with the difference similarly falling with increasing exponent $X$).
In other words, the threshold of $0.5$ used in target prediction introduces another regressive effect that moves the predicted target rate for a given group below the observed target rate: if the target rate is low, it is less likely for a leaf to contain enough examples of the target to reach the $0.5$ prediction threshold, producing systematic underprediction of the target. If the target rate is high, by contrast, this threshold effect results in overprediction of the target. In the case of decision trees (or any decision-threshold ML algorithm), we expect that both the SSIE and the decision threshold effects will contribute to produce systematic patterns of underprediction.

\subsection{Predicting bias}
Given the above results and theory, we can now make a prediction regarding the effect of SSIE and decision threshold bias from the structure of the data. 
Both types of bias will be amplified for inferences where the target rate significantly different from 0.5 (much higher or lower). This is due to the regressive effects of DI and the decision threshold acting away from 0.5 and getting weaker the closer the target rate is 0.5.
We also expect the ES to increase SSIE bias (i.e. if a large proportion of the relevant PVCs/leaves in a group are small, then it will be subject to greater bias). Generally, the minority group is more likely to have smaller leaf sizes and so will be subject to greater bias, but this is not always the case. 
In cases where the above points do not hold (target rate is close to 0.5 or there are few small leaves), then SSIE bias will not have a significant effect and other forms of bias are more likely influencing the predictions. However, when applying algorithms that use a decision threshold, we will always see this underprediction effect as explained in the previous section. Therefore, our theory gives us a method of explaining two significant sources of bias that we see in predictions. 

\subsection{Example from the Literature}

We now examine an example of ML bias from the literature. \cite{cunningham2020} explores ML predictions on the adult dataset which contains majority and minority groups, a rare target trait and other predictor variables. In it, women were underrepresented in the $>50k$ Target category compared to men (see Table \ref{tab:adult_data}). A random forest model's predictions increased this underrepresentation (actual: 11\% of women and 30\% of men in the $>50k$ group, predicted: 8\% of women and 26\% of men in the $>50k$ group) with the effect more pronounced for women than men (a decline of 25\% relative to the actual values for women, but 15\% for men; see Table \ref{tab:F_model_predictions}). The results in Table \ref{tab:F_model_predictions} show the model's predictions.

\begin{table}[H]
\centering
\caption{\ref{tab:adult_data} Proportion Table from `Adult' Data showing target rate for each group in bold}
\label{tab:adult_data}
\begin{tabular}{l|ll}
           & Male & Female \\ \hline
Target=0          & 0.47 & 0.29  \\
Target=1          & 0.20 (\textbf{30\%}) & 0.04 (\textbf{12\%})   \\ 
\end{tabular}
\end{table}

\begin{table}[H]
\centering
\caption{Proportion Table of the Random Forest Predictions, with percentage rise or fall relative to sample proportions in Table \ref{tab:adult_data}}
\label{tab:F_model_predictions}
\begin{tabular}{l|ll}
           & Male & Female \\ \hline
Target=0          & 0.50 & 0.30     \\
Target=1          & 0.17 (\textbf{25\%}; 17\% fall) & 0.03 (\textbf{9\%}; 25\% fall)   \\
\end{tabular}
\end{table}
 
As before, there is underprediction for both groups, but a greater underprediction for the minority. 

We can rerun the example from section 3.1 using the values from the adult data set (while still using the single PVC case) to get the results seen in Table \ref{tab:pvx}.

\begin{table}[H]
\centering
\caption{Proportion table for DI from samples, with percentage rise or fall relative to sample proportions in Table \ref{tab:pv2}}
\label{tab:pvx}
\begin{tabular}{l|ll}
           & Maj & Min \\ \hline
Target=0          & 0.48  & 0.30   \\
Target=1          & 0.19 (\textbf{29\%}; 3\% fall) & 0.03 (\textbf{9\%}; 25\% fall)   \\
\end{tabular}
\end{table}

As predicted above, the single PVC case sufficiently describes the process of ML inference. The infrequent feature combinations dominate the inference and their small sample size contributes significant bias to the predictions.

\section{Results}

In order to provide greater evidence for our theory, we explored the patterns of bias for various subsets of the `adult' and COMPAS datasets \cite{adultkohavi, uciadult, compas2016}. Regarding pre-processing, we took generally standard approaches to convert the variables to categorical and then one hot encoded to binary (mostly following the approach of \cite{quy2021survey} for `adult' and \cite{compas2016} for COMPAS). In addition, the `race' variable was converted to `white=0' or `white=1' for `adult' and `nonwhite=0' or `nonwhite=1' for COMPAS (as the `nonwhite' group was in the majority in COMPAS).

Once pre-processed, we examined each possible subset in the dataset based on the majority and minority of the split (e.g. after one hot encoding, individuals with `Never-married'$=0$, those with `Never-married'$=1$, 'Local-gov'$=0$, etc.). The splits were filtered so that the minority group had at least $100$ members and that each group had target occurrences. This was to ensure the inference scenarios being tested were realistic and not influenced by the random variation of small datasets. For each subset, we examined the predictions of an sklearn decision tree model \cite{sklearn}. This gave us the difference between observed target rate and predicted rate for each subset (the `bias'):

\begin{equation}
b(D) = \frac{pred - act}{act}
\end{equation}

where $D$ is the subset being considered, $pred$ is the predicted target rate and $act$ is the actual target rate.

Given our theory, we have several different approaches for predicting this bias from the data.
The first method would be to examine the target rate of the subset. Given the regressive effects of the SSIE, a target rate of less than $0.5$ will result in underprediction of the target rate with the effect increasing as the target rate approaches $0$. A target rate of greater than $0.5$ will result in overprediction of the target rate with the effect increasing as the target rate approaches $1$. 
Another approach would be to consider the relationship between ES (Equation. \ref{eq:ES}) and the bias. A larger number of infrequent PVCs will contribute more bias to the overall predicted target rates.
Yet another approach would be the association between some combination of the target rate and ES and the bias. Equation \ref{eq:trES} is a modification of the ES expression that adds the target rate at each sum:

\begin{equation}
    \label{eq:trES}
    ES = \sum_{i} \frac{p_{i} + Tr}{i}
\end{equation}
where Tr is the target rate in the sample.

The resulting correlations for the majority and minority subsets of `adult' and COMPAS are summarised in tables \ref{tab:majminres} and \ref{tab:fullres}. `Tr' is the correlation between that groups target rate and the observed underprediciton while `Tr+ES' is the correlation between the sum of the target rate and ES with the observed underprediction. `Diff' considers the difference in underprediction between groups and `full' combines the results of all subsets.
We see that strong correlations exist between the target rate and observed underprediction in all scenarios. Low target rates will regress towards $0$ and high target rates towards $1$ due to the SSIE. 
The addition of ES improves this correlation for the minority groups in `adult' and the `full' `adult' data. This indicates that in these smaller groups, the high levels of ES are contributing to the overall bias. `adult' contains many more variables, and therefore many more PVCs, than COMPAS, so we would expect ES to play a larger role. 
In general, the overall target rate is sufficient to estimate the expected bias due to SSIE for a given dataset with ES being a helpful add on. 
Given the many splits considered, and their diversity in size and target rate, this is significant evidence to support our theory of the SSIE.

\begin{table}[H]
\centering
\caption{Correlations between Underprediction and Predicted Underpredictions for Majority and Minority Groups}
\label{tab:majminres}
\begin{tabular}{r|lrrrr}
    Data   &   Maj Tr &   Maj Es+Tr &   Min Tr &   Min Es+Tr \\
\hline
adult  &     0.82 &        0.82 &     0.49 &        0.56 \\
COMPAS &     0.95 &        0.95 &     0.86 &        0.86 \\
\end{tabular}
\end{table}

\begin{table}[H]
\centering
\caption{Correlations between Underprediction and Predicted Underpredictions for Differences and Full Datasets}
\label{tab:fullres}
\begin{tabular}{r|lrrr}
    Data   &   Diff Es+Tr &   Full Tr &   Full Es+Tr \\
\hline
adult  &         0.61 &      0.5  &         0.56 \\
COMPAS &         0.47 &      0.86 &         0.85 \\
\end{tabular}

\end{table}

Obviously, this bias due to SSIE is one of many sources of bias in an ML application. Extensive analysis of bias due to other sources (societal bias, other statistical bias, methodology bias, etc.) can be found in \cite{mehrabibias, olteanu, suresh}. In particular, when the prerequisites for this SSIE (as described in section 3.3) are not met, and bias is still observed in the model's predictions, we expect these other sources are significantly contributing. \newline
Our analysis of the COMPAS data initially showed agreement with our theory: greater underprediction for the minority (`white') group than the majority (`non-white'). However, given the evidence proposed in \cite{compas2016} that this bias is due to societal factors embedded in the data, we conclude that the SSIE is not a major contributor. 

\section{Discussion}

An alternative perspective of SSIE bias is that it is a bias of sampling, not inference. Given the demonstrable negative effects on minority groups, the `correct' approach would be to not conduct inference on small samples in the first place. If all samples were large enough, no bias would be observed when applying DI. 
One may suggest that increasing data collection for this group may solve this issue. However, simply increasing the sample size in absolute terms will not reduce the SSIE bias, notice that we also see this bias in the larger majority group, albeit to a lesser extent. To truly `eliminate' this bias, the data generation process would have to ensure that every PVC in the data occurs more than a set number of times. Given the thousands of possible PVCs in most ML datasets, this quickly becomes infeasible through extra sampling alone. Instead of increasing the sample size, an easier approach would be to reduce the number of possible PVCs by removing predictor variables from the data. This introduces an `information/bias tradeoff', impacting model performance. Further investigation of this solution to SSIE bias is required to determine if the effectiveness of this approach.

\section{Conclusion}
ML bias is a pervasive issue with serious social impacts. We have shown that theoretical models of DI and decision thresholds accurately depict results from the ML literature. Our theory describes the SSIE as baseline of unavoidable bias, inherent to rational decision making, solely due to group size within the sample. In addition, we have demonstrated how decision thresholds introduce a similar bias. These biases effects members of minority groups more severely and are present even in large datasets. Further research is required to explore this effect for other ML model types, such as more advanced tree based models and neural networks.

\bibliography{mybibfile}

\end{document}